\documentclass[letterpaper, 10 pt, conference]{ieeeconf}  

\IEEEoverridecommandlockouts                              

\usepackage{graphics} 
\usepackage{epsfig} 
\usepackage{mathptmx} 
\usepackage{times} 
\usepackage{amsmath} 
\usepackage{amssymb}  
\usepackage{xcolor}
\usepackage{subcaption} 

\renewenvironment{align*}{\align}{\endalign}

\title{\LARGE \bf
EndoSensorFusion:  Particle Filtering-Based Multi-sensory Data Fusion with Switching State-Space Model for Endoscopic Capsule Robots
}

\author{Mehmet Turan$^{1}$, Yasin Almalioglu$^{2}$, Hunter Gilbert$^{3}$, Helder Araujo$^{4}$, Taylan Cemgil$^{5}$, Metin Sitti$^{6}$
\thanks{$^{1}$Mehmet Turan is with Physical Intelligence Department, Max Planck Institute for Intelligent Systems, Germany and Department of Information Technology and Electrical Engineering, ETH Zurich, Switzerland 
        {\tt\small mturan@student.ethz.ch}}%
\thanks{$^{2}$Yasin Almalioglu is with Computer Engineering Department, Bogazici Univesity, Turkey
        {\tt\small yasin.almalioglu@boun.edu.tr}}%
\thanks{$^{3}$Hunter Gilbert is with the Mechanical Engineering Department, Louisiana State University, Baton Rouge, LA 70803, USA
        {\tt\small hbgilbert@lsu.edu}}%
\thanks{$^{4}$Helder Araujo is with Institute for Systems and Robotics, University of Coimbra, Portugal
        {\tt\small helder@isr.uc.pt}}%
\thanks{$^{5}$Taylan Cemgil is with Computer Engineering Department, Bogazici Univesity, Turkey
        {\tt\small taylan.cemgil@boun.edu.tr}}%
\thanks{$^{6}$Metin Sitti is with Physical Intelligence Department, Max Planck Institute for Intelligent Systems, Germany
        {\tt\small sitti@is.mpg.de}}%
}

\begin{document}

\maketitle
\thispagestyle{empty}
\pagestyle{empty}

\begin{abstract}
A reliable, real time multi-sensor fusion functionality is crucial for localization of actively controlled capsule endoscopy robots, which are an emerging, minimally invasive diagnostic and therapeutic technology for the gastrointestinal (GI) tract. 
In this study, we propose a novel multi-sensor fusion approach based on a particle filter that incorporates an on-line estimation of sensor reliability and a non-linear kinematic model learned by a recurrent neural network. 
Our method sequentially estimates the true robot pose from noisy pose observations delivered by multiple sensors. 
We experimentally test the method using 5 degree-of-freedom (5-DoF) absolute pose measurement by a magnetic localization system and a 6-DoF relative pose measurement by visual odometry. 
In addition, the proposed method is capable of detecting and handling sensor failures by ignoring corrupted data, providing the robustness expected of a medical device.
Detailed analyses and evaluations are presented using ex-vivo experiments on a porcine stomach model prove that our system achieves high translational and rotational accuracies for different types of endoscopic capsule robot trajectories.	
\end{abstract}

\section{INTRODUCTION}

One of the highest potential scientific and social impacts of milli-scale, untethered, mobile robots is their healthcare applications. 
Swallowable capsule endoscopes with an on-board camera and wireless image transmission device have been commercialized and used in hospitals (FDA approved) since 2001, which has enabled access to regions of the GI tract that were impossible to access before, and has reduced the discomfort and sedation related work loss issues \cite{IddanWirelessNature00, sitti2015biomedical,  turan2017non, turan2017deep}. 
However, with systems commercially available today, capsule endoscopy cannot provide precise (centimeter to millimeter accurate) localization of diseased areas, and active, wireless control remains a highly active area of research.
Several groups have recently proposed active, remotely controllable robotic capsule endoscope prototypes equipped with additional functionalities such as localized drug delivery, biopsy and other medical functions \cite{munoz2014review, carpi2011magnetically, yim2014biopsy, petruska2013omnidirectional, son20165, son2017magnetically, turan2017sparse, turan2017fully, turan2017endovo}. Accurate and robust localization would not only provide better diagnostic information in passive devices, but would also improve the reliability and safety of active control strategies like remote magnetic actuation.

In the last decade, many different approaches have been developed for real time endoscopic capsule robot localization, including received signal strength (RSS), time of flight and time difference of arrival (ToF and TDoA), angel of arrival (AoA) and radiofrequency (RF) identification (RFID)-based methods \cite{munoz2014review, carpi2011magnetically}. Recently, it has also been shown that the permanent magnets added to capsule robots for remote magnetic actuation can be simultaneously used for precision localization \cite{son20165}. 
This strategy has a clear advantage for miniaturization: the permanent magnet provides two essential functions rather than one.

Hybrid techniques based on the combination of different measurements can improve both the accuracy and the reliability of the location measurement system.
Sensor fusion techniques have been applied to wireless capsule endoscopes, and several combinations of sensor types have been investigated.
The first subgroup of hybrid techniques fuses radio frequency (RF) signals and video for localization of the capsule robot \cite{bao2015hybrid, geng2016design}. Geng et al.\ assert that using both RF signals and video data can result in millimetric accuracy while previous techniques were able to achieve only a few centimeters accuracy. 
One drawback of these studies is that most techniques for data fusion have been based on Kalman filtering, which works best for linear systems, while the kinematics and dynamics of capsule robots are nonlinear in orientation.

In the second group, RF signal and magnetic localization are fused for the localization of the capsule robot \cite{umay2016adaptive, umay2017adaptive}. 
In these studies, a localization method that has high accuracy for simultaneous position and orientation estimation has been investigated. 
In the third group of hybrid techniques, video and magnetic localization are fused \cite{gumprecht2013navigation}. 
In \cite{gumprecht2013navigation}, the authors introduced an ultrasound imaging-based localization combined with magnetic field-based localization.
\par

Although some of these state-of-the-art sensor fusion techniques have achieved remarkable accuracy for the tracking and localization task of a capsule robot, they are not able to detect and autonomously handle sensor faults, and additionally several techniques using RF localization require complex signal corrections to account for attenuation and propagation of RF signals inside human body tissues.
In addition, most previous models use relatively simple dynamic models for the capsule, whereas performance would be greatly improved by a more accurate model of the system. 
Lastly, previously demonstrated methods generate inaccurate estimations in cases where noise from the environment and the actuation system interferes with one or more components of the localization system.

In this paper, we propose a novel multi-sensor fusion algorithm for capsule robots based on switching state space models with particle filtering using the endoscopic capsule robot dynamics modelled by Recurrent Neural Networks (RNNs), which can handle sensor faults and non-linear motion models. 
The main contributions of our paper are as follows: 

\begin{itemize}
	\item To the best of our knowledge, this is the first multi-sensor data fusion approach that combines a switching observation model, a particle filter approach, and a recurrent neural network developed for the endoscopic capsule robot and hand-held  endoscope localization.
	\item We propose a sensor failure detection system for endoscopic capsule robots based on probabilistic graphical models with efficient proposal distributions applied onto the particle filtering. The approach can be generalized to any number of sensors and any mobile robotic platforms.
	\item No manual formulation is required to determine a probability density function that describes the motion dynamics, contrary to traditional particle filter and Kalman filter based methods. 
\end{itemize}

The paper is organized as follows.
Section~\ref{sec:fusion} introduces the sensor fusion algorithms and combination with the RNN-based dynamic model.
Section~\ref{sec:experimental_method} describes the experiments used to verify the proposed methods for a wireless capsule endoscope in an ex-vivo porcine model.
Section~\ref{sec:results_and_discussion} includes the results and discussion of the experiments, and section \ref{sec:conclusion} concludes with future directions.

\section{Sensor Fusion and Modeling Approach}
\label{sec:fusion}
The particle filter is a statistical Bayesian filtering method to compute the posterior probability density functions (pdf) of sequentially obtained state vectors $\mathbf{x_{t}} \in \mathcal{X}$ which are suggested by (complete or partial) sensor measurements. 
For the capsule robot, the state $\mathbf{x_{t}}$ is composed of the 6-DoF pose, which is assumed to propagate in time according to a model:

\begin{equation}
\label{eq:state_transition}
\mathbf{x}_{t} = f(\mathbf{x}_{t-1} , \mathbf{v}_{t} )  
\end{equation}
where $f$ is a non-linear state transition function and $\mathbf{v_{t}}$ is a white noise distributed. 
$t$ is the index of a time sequence, $t \in \{1,2,3, ...\}$. 

6-DoF pose state estimation with a high precision is a complex problem, which often requires multi-sensor input or sequential observations. 
In our capsule, we have two sensor systems, one being a 5-DoF magnetic sensor array and the other one being an endoscopic monocular RGB camera (these subsystems are described later). 
Generally speaking, observations of the pose are produced by $n$ sensors $\mathbf{z}_{k,t}  (k = 1,...,n)$,  where the probability distribution $p(\mathbf{z}_{k,t}|\mathbf{x}_t)$ is known for each sensor.

\begin{figure}
      \centering
      \includegraphics[width=0.9\linewidth]{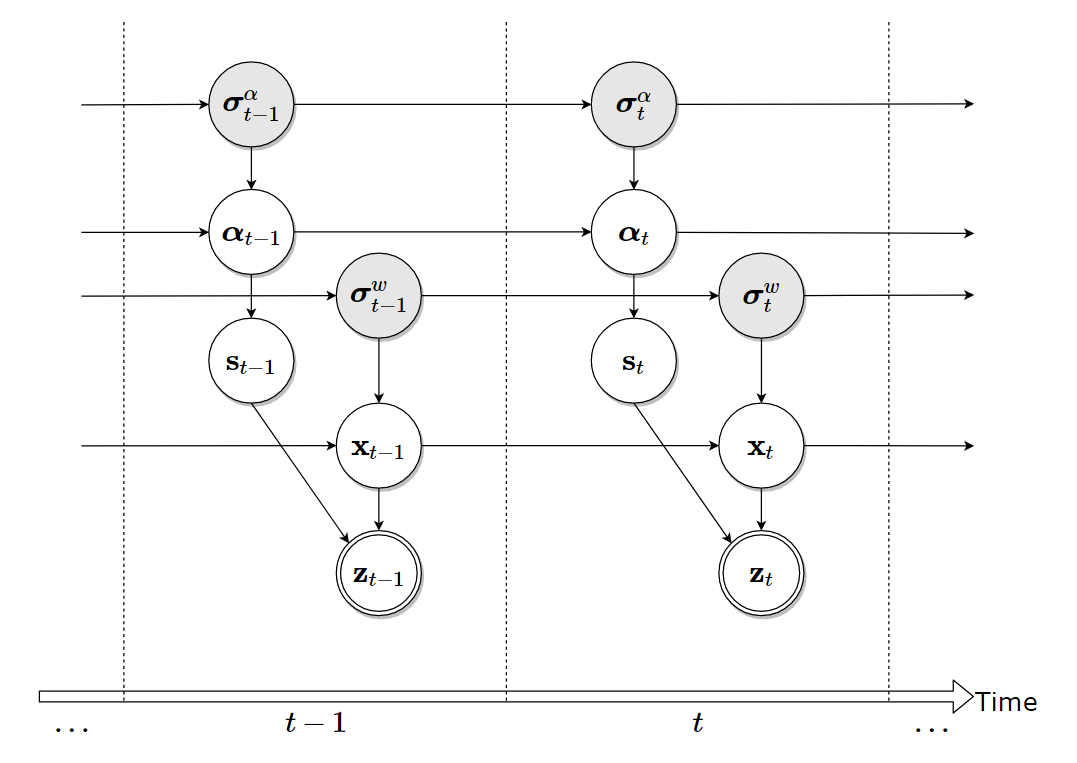}
      \caption{The overall switching state-space model. The double circles denote observable variables and the gray circles denote hyper-parameters.}
      \label{fig:graph_model}
\end{figure}

\subsection{The Sequential Bayesian Model and Problem Statement}
\label{sec:model}

We estimate the 6-DoF pose states which rely on latent (hidden) variables by using the Bayesian filtering approach. 
The probabilistic graphical model that shows the relations between all of the variables is shown in Fig.~\ref{fig:graph_model}. The hidden variables of sensor states are denoted as $s_{k,t}$, which we call switch variables, where $s_{k,t}\in \{0,..., d_{k}\}$ for $k = 1,...,n$. $d_{k}$ is the number of possible observation models, e.g., failure and nominal sensor states. 
The observation model for $\mathbf{z}_{k,t}$ can be described as:

\begin{equation}
\mathbf{z}_{k,t} = h_{k, s_{k, t},t}(\mathbf{x}_{t}) + \mathbf{w}_{k, s_{k, t},t} 
\end{equation}
where $h_{k, s_{k, t},t}(\mathbf{x}_{t})$ is the non-linear observation function and $\mathbf{w}_{k, s_{k, t},t} $ is the observation noise. The latent variable of the switch parameter $s_{k, t}$ is defined to be $0$ if the sensor is in a failure state, which means that observation $\mathbf{z}_{k,t}$ is independent of $\mathbf{x}_{t}$, and $1$ if the sensor $k$ is in its nominal state of work. The prior probability for the switch parameter $s_{k,t}$ being in a given state $j$, is denoted as  $\alpha_{k,j,t}$ and it is the probability for each sensor to be in a given state:

\begin{equation}
Pr(s_{k,t}=j) = \alpha_{k,j,t} , \quad 0 \leq j \leq d_{k} 
\end{equation}
where $\alpha_{k,j,t} \geq 0$ and  $\sum_{j=0}^{d_{k}}\alpha_{k,j,t}=1$ with a Markov evolution model. The objective posterior pdf $p(\mathbf{x}_{0:t}, \mathbf{s}_{1:t}, \mathbf{\alpha}_{0:t}|\mathbf{z}_{1:t})$ and the marginal posterior probability $p(\mathbf{x}_{t}|\mathbf{z}_{1:t})$ , in general, cannot be determined in a closed form due to its complex shape. However, sequential Monte Carlo methods (\textit{particle filters}) provide a numerical approximation of the posterior pdf with a set of samples (\textit{particles}) weighted by the kinematics and observation models.

\subsection{Proposal Distributions}
In this section, we formulate the optimal proposal distributions in terms of minimizing the variance of the weights and effective approximations, in cases where sampling from the optimal distributions is not feasible. The particles are extended from time $t-1$ to time $t$ according to the importance distribution denoted by $q(\cdot)$.

\begin{itemize}
\item
$q(\mathbf{x}_t \mid \mathbf{x}_{t-1}^{(i)}, {\mathbf{\sigma}_t^w}^{(i)}, \mathbf{\hat{s}}_t^{(i)}, \mathbf{z}_t)$ is approximated by an unscented Kalman filter (UKF) step:
$$\hat{\mathbf{x}}_{t|t}^{(i)} = \hat{\mathbf{x}}_{t|t-1}^{(i)} + \sum_{k=1}^{n}\hat{s}_{k,t}^{(i)} K_{k,t}^{(i)} \hat{\mathbf{\nu}}_{k,t}^{(i)}$$ where $\hat{\mathbf{x}}_{t|t-1}^{(i)}=f(\mathbf{x}_{t-1}^{(i)})$, $n$ is the number of sensors, $\hat{\mathbf{\nu}}_{k,t}^{(i)}$ is the residual, and $K_{k,t}^{(i)}$ is the Kalman gain sequentially obtained by UKF. Finally,
$$q(\mathbf{x}_t \mid \mathbf{x}_{t-1}^{(i)}, {\mathbf{\sigma}_t^w}^{(i)}, \mathbf{\hat{s}}_t^{(i)}, \mathbf{z}_t) = \mathcal{N}(\mathbf{x}_t; \hat{\mathbf{x}}_{t|t}^{(i)}, P_{t|t}^{(i)})$$
where the error covariance matrix, $P_{t|t}^i$ is obtained by the UKF step with the process noise of ${\mathbf{\sigma}_t^w}^{(i)}$.

\item In switching state-space models, the switch parameters with self-adaptive prior are more efficient   than a fixed prior approach \cite{caron2007particle, hue2002sequential}. The optimal proposal distribution for switch variable that represents the state of a sensor is given by

\begin{equation}
Pr(\mathbf{s}_{k,t}|\mathbf{x}^{(i)}_{t-1}, \mathbf{\alpha}^{(i)}_{k, t-1}, \mathbf{z}_{k, t}) = \frac{\alpha^{(i)}_{k, s_{k, t}, t-1} p(\mathbf{z}_{k,t}| s_{k,t},\mathbf{x}^{(i)}_{t-1})}{\sum_{j=0}^{d_{k}}\alpha^{(i)}_{k, s_{k, t}, t-1} p(\mathbf{z}_{k,t}|j,\mathbf{x}^{(i)}_{t-1})}
\end{equation}
which is approximated by applying UKF to pdfs $p(\mathbf{z}_{k,t}|j,\mathbf{x}^{(i)}_{t-1})$ for $j=0,...,d_{k}$ 

\begin{equation}
p(\mathbf{z}_{k,t}|j,\mathbf{x}^{(i)}_{t-1}) \simeq \mathcal{N}(h_{k,j,t}(\hat{x}^{(i)}_{t|t-1}), S^{(i)}_{k, j, t})
\end{equation}
where $\hat{x}^{(i)}_{t|t-1} = f(\mathbf{x}^{(i)}_{t-1})$ is the state prediction and $S^{(i)}_{k, j, t}$ is the approximated innovation covariance matrix approximated by UKF. Hence, the proposal distribution for the switch parameter $\mathbf{s}_{k,t}$ is given by

\begin{equation}
\begin{split}
q\big(\mathbf{s}_{k,t}|\mathbf{x}^{(i)}_{t-1}, \mathbf{\alpha}^{(i)}_{k, t-1}, \mathbf{z}_{k, t}\big) \\ \propto \alpha^{(i)}_{k, s_{k}, t-1} \mathcal{N}(h_{k, s_{k}, t-1}(\hat{x}^{(i)}_{t|t-1}), S^{(i)}_{k, s_{k}, t})
\end{split}
\end{equation}

\item
The optimal proposal distribution for the hyperparameter $\sigma^{\alpha}_{k, t-1}$ is calculated in closed form as 
\begin{equation}
\begin{split}
q&\bigg(log(\sigma^{\alpha}_{k, t})|\mathbf{\alpha}^{(i)}_{k, t}, \mathbf{\alpha}^{(i)}_{k, t-1}, \sigma^{\alpha(i)}_{k, t-1}\bigg)\\
&= \frac{D\bigg(\mathbf{\alpha}^{(i)}_{k, t}; \sigma^{\alpha}_{k, t}\mathbf{\alpha}^{(i)}_{k, t-1} \bigg)}{D\bigg(\mathbf{\alpha}^{(i)}_{k, t}; \sigma^{\alpha}_{k, t-1}\mathbf{\alpha}^{(i)}_{k, t-1} \bigg)}\\
&\times \mathcal{N}\bigg(log(\sigma^{\alpha}_{k, t}); log(\sigma^{\alpha(i)}_{k, t-1}), \lambda^{\alpha}\bigg).
\end{split}
\end{equation}

We generate samples from the distribution with Adaptive Rejection Sampling (ARS) method since direct sampling is not feasible \cite{gilks1992adaptive}. Using ARS, the need for locating the supremum diminishes because the distribution is log-concave. Another advantage of ARS is that it uses recently acquired information to update the envelope and squeezing functions, which reduces the need to evaluate the distribution after each rejection step. Fig.~\ref{fig:ars} shows an ARS sampling result indicating the effectiveness of the applied sampling method for the proposal distribution. It can be seen in Fig.~\ref{fig:ars} that a tight piecewise hull has converged to the target distribution after rejection steps and interior knots are regenerated in the vicinity of the expected values. 

\begin{figure}
      \centering
      \includegraphics[width=0.9\linewidth, height=0.9\linewidth]{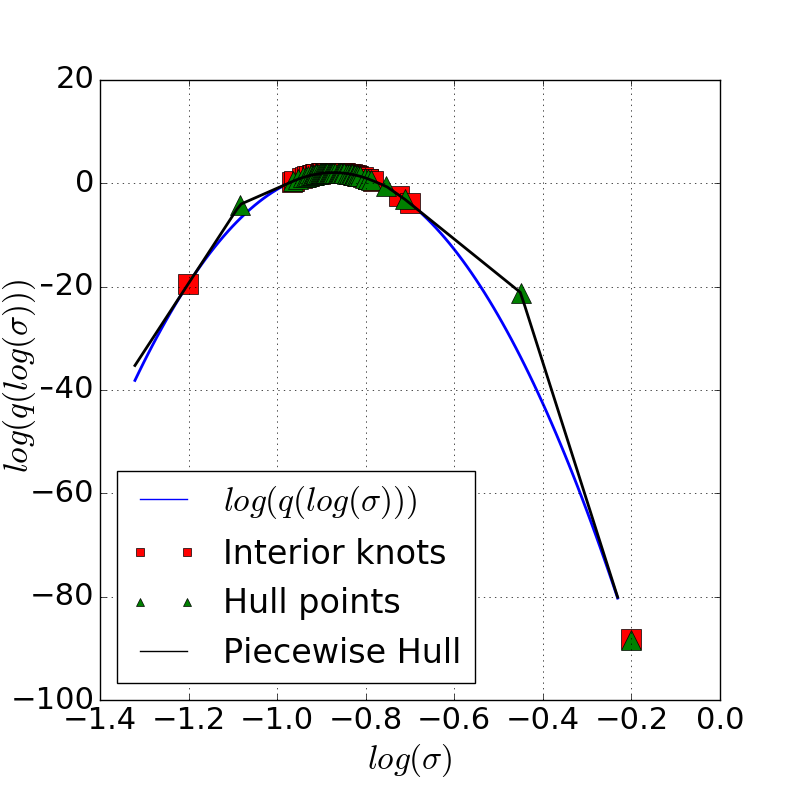}
      \caption{Example ARS sampling result for $log(\sigma_{k,t})$. The piecewise hull and the generated samples are shown.}
      \label{fig:ars}
   \end{figure}

\item Considering that the Dirichlet distribution is conjugate to the multinomial distribution, the optimal proposal distribution for the confidence parameter $\mathbf{\alpha}_{k,t}$ can be reformulated in a closed form as a Dirichlet distribution with a decreasing variance parameter for failure sensor states. 

\end{itemize}

\begin{figure}
\centering
\includegraphics[width=\columnwidth]{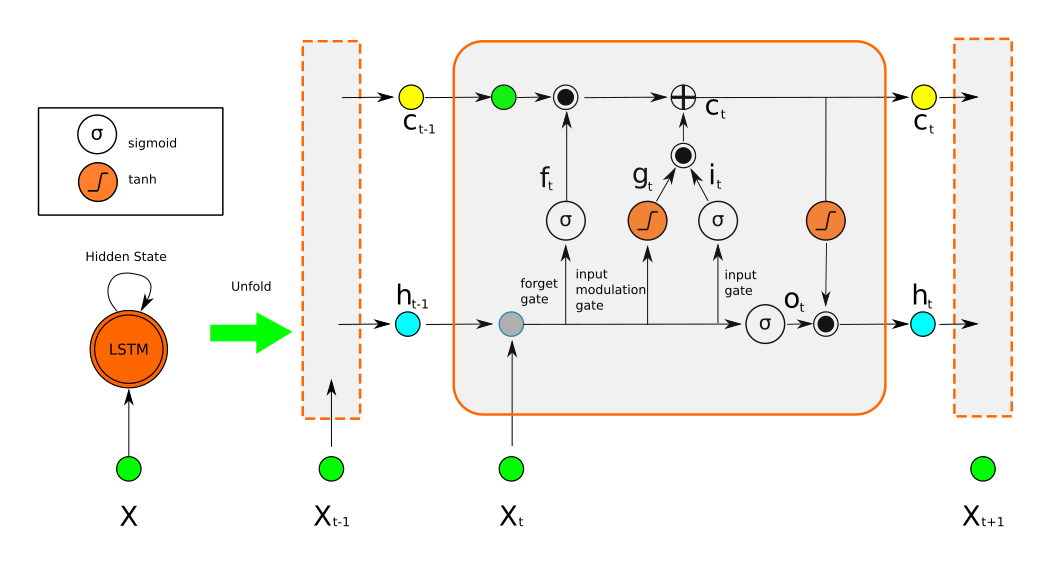}
\caption{Information flow through the units of the LSTM \cite{hochreiter1997long}}
\label{fig:rnn}
\end{figure}

\begin{figure}
	\includegraphics[width=\columnwidth]{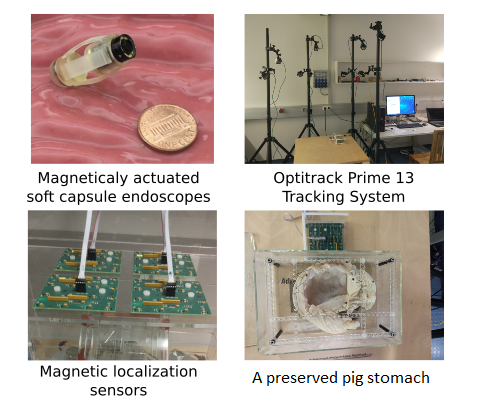}
	\caption{Experimental setup}
	\label{fig:architecture}       
\end{figure}

\subsection{RNN-based Kinematics Model} 
Existing sensor fusion methods based on traditional particle filter and Kalman filter approaches have their limitations when applied to nonlinear dynamic systems. 
The Kalman filter and extended Kalman filter assume that the underlying dynamic process  is well-modeled by linear equations or that these equations can be linearised without a major loss of fidelity. 
On the other hand, particle filters accommodate a wide variety of dynamic models, allowing for highly complex dynamics in the state variables. 

In the last few years, deep learning (DL) techniques have provided solutions to many computer vision and machine learning tasks.
Contrary to these high-level tasks, multi-sensory data fusion is mainly working on motion dynamics and relations across sequence of pose observations obtained from sensors, which can be formulated as a sequential learning problem. 
Unlike traditional feed-forward artificial neural networks, RNNs are very suitable for modelling the dependencies across time sequences and for creating a temporal motion model since it has a  memory of hidden states over time and has directed cycles among hidden units, enabling the current hidden state to be a function of arbitrary sequences of inputs. 
Thus, using an RNN, the pose estimation of the current time step benefits from information encapsulated in previous time steps \cite{walch2016image} and is suitable to formulate the state transition function $f$ in Equation \ref{eq:state_transition}.
In particular, information about the most recent velocities and accelerations become available to the model.

Long Short-Term Memory (LSTM) is a suitable implementation of RNN to exploit longer trajectories since it avoids the vanishing gradient problem of RNN resulting in a higher capacity of learning long-term relations among the sequences by introducing memory gates such as input, forget and output gates, and hidden units of several blocks. The information flow of the LSTM is shown in Fig.\ref{fig:rnn}.
The input gate controls the amount of new information flowing into the current state, the forget gate adjusts the amount of existing information that remains in the memory, and the output gate decides which part of the information triggers the activations. Given the input vector $x_{k}$ at time $k$, the output vector $h_{k-1}$ and the cell state vector $c_{k-1}$ of the previous LSTM unit, the LSTM updates at time step $k$ according to the following equations:
\begin{align*}
f_k &= \sigma(W_{f}\cdot [x_k, h_{k-1}] + b_f) \\
i_k &= \sigma(W_{i}\cdot [x_k, h_{k-1}] + b_i) \\
g_k &= \mathrm{tanh}(W_{g}\cdot[x_k, h_{k-1}] + b_g) \\
c_k &= f_k\odot c_{k-1} + i_k\odot g_k \\
o_k &= \sigma(W_{o}\cdot[x_k, h_{k-1}] + b_o) \\
h_k &= o_k\odot \mathrm{tanh}(c_k)
\end{align*}
where $\sigma$ is sigmoid non-linearity, tanh is hyperbolic tangent non-linearity, $W$ terms denote corresponding weight matrices, $b$ terms denote bias vectors, $i_{k}$, $f_{k}$, $g_{k}$, $c_{k}$ and $o_{k}$ are input gate, forget gate, input modulation gate, the cell state and output gate at time $k$, respectively, and $\odot$ is the Hadamard product \cite{gers1999learning}.

\section{Experimental Setup and Dataset} \label{sec:experimental_method}

\subsection{Magnetically Actuated Soft Capsule Endoscopes (MASCE)}
\begin{figure}[t!]
\centering
	\includegraphics[width=\columnwidth]{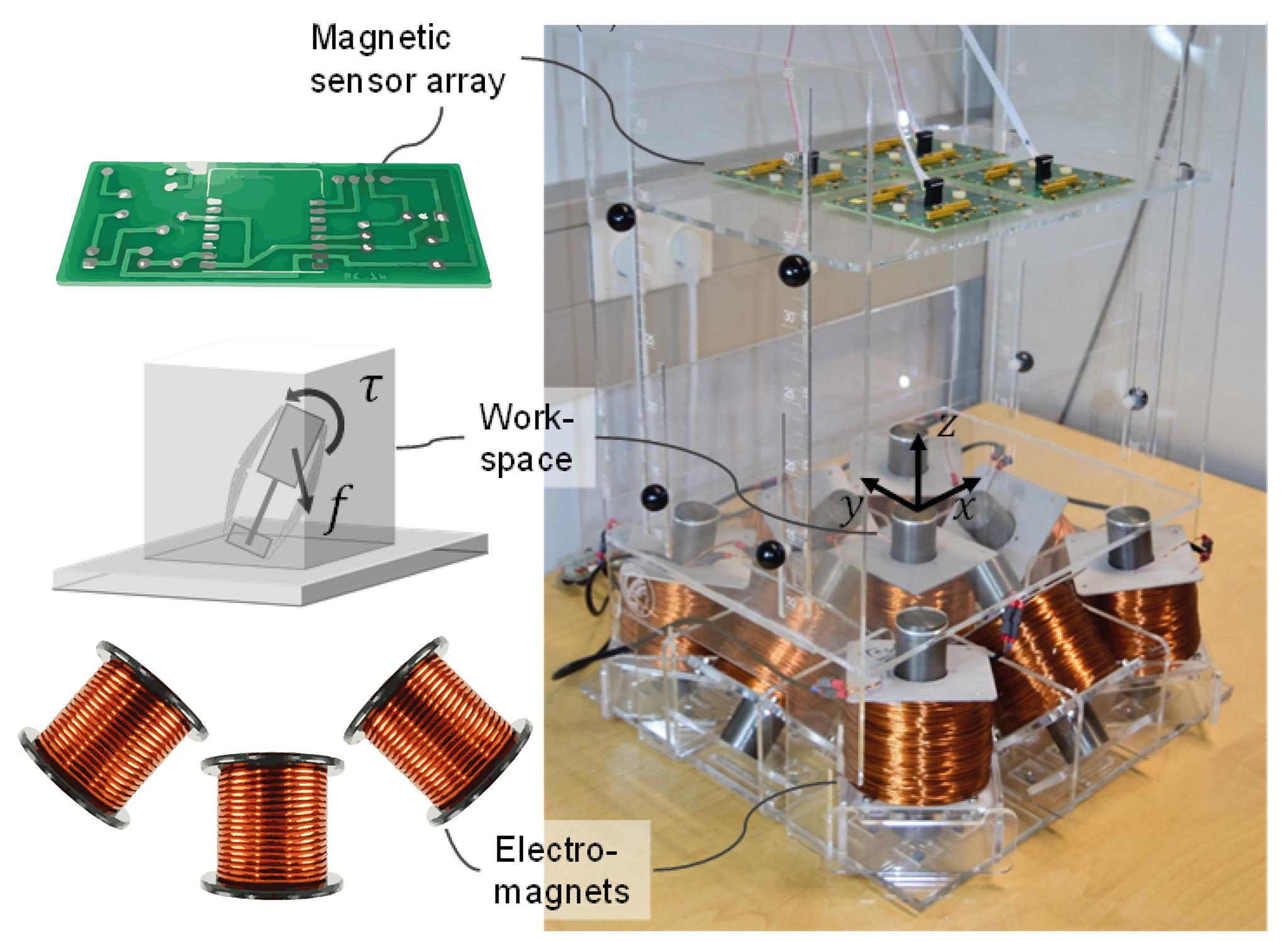}
\caption{Actuation system of the MASCE \cite{son2017magnetically}}
\label{fig:capsule_setup}
\end{figure}

Our capsule prototype is a magnetically actuated soft capsule endoscope (MASCE) designed for disease detection, drug delivery and biopsy operations in the upper GI-tract. The prototype is composed of a RGB camera, a permanent magnet, an empty space for drug chamber and a biopsy tool (see Figs.~\ref{fig:architecture} and \ref{fig:capsule_setup} for visual reference). The magnet exerts magnetic force and torque to the robot in response to a controlled external magnetic field \cite{son2017magnetically}. The magnetic torque and forces are used to actuate the capsule robot and to release drug. Magnetic fields from the electromagnets generate the magnetic force and torque on the magnet inside MASCE so that the robot moves inside the workspace. Sixty-four three-axis magnetic sensors are placed on the top, and nine electromagnets are placed in the bottom \cite{son2017magnetically}.

\subsection{Magnetic Localization System}
Our 5-DoF magnetic localization system is designed for the position and orientation estimation of untethered meso-scale magnetic robots \cite{son20165}. The system uses external magnetic sensor system and electromagnets for the localization of the magnetic capsule robot. A 2D-Hall-effect sensor array measures the component of the magnetic field from the permanent magnet inside the capsule robot at several locations outside of the robotic workspace. Additionally, a computer-controlled magnetic coil array consisting of nine electromagnets generates actuator's magnetic field. The core idea of our localization technique is separation of capsule's magnetic field from actuator's magnetic field. For that purpose, actuator's magnetic field is subtracted from the magnetic field data which is acquired by Hall-effect sensor array. As a further step, second-order directional differentiation is applied to reduce the localization error \cite{son20165}. 

\subsection{Monocular Visual Odometry} 
The visual odometry is performed by minimization of a multi-objective cost function including alignment of sparse features, photometric and volumetric correlation. For every input RGB image, we create its depth image using the source code of the perspective shape-from-shading under realistic lighting conditions project \cite{visentini2012metric, ciuti2012intra}. Once depth map is obtained, the framework uses both RGB and depth map information to jointly estimate camera pose. An energy minimization based  pose estimation technique is applied containing both sparse optical flow (OF) based correspondence establishment, and volumetric and photometric dense alignment establishment \cite{dai2017bundlefusion, whelan2015elasticfusion, newcombe2011kinectfusion}.  Inspired from the pose estimation strategies proposed by \cite{dai2017bundlefusion, whelan2015elasticfusion, newcombe2011kinectfusion}, for a parameter vector 
\begin{equation}
X=(R_o,t_o,...,R_\textrm{$|S|$} ,t_\textrm{$|S|$})^T 
\end{equation}
for $|S|$ frames, the alignment problem is defined as a variational non-linear least squares minimization problem with the following objective, consisting of the OF based pixel correspondences and dense jointly photometric-geometric constraints\cite{dai2017bundlefusion, whelan2015elasticfusion, newcombe2011kinectfusion}. Outliers after OF estimation are eliminated using motion bounds criteria, which removes pixels with a very large displacement and too different motion vector than neighbouring pixels. The energy minimization equation is as follows:
\begin{equation}
\label{eq:main}
E_\textrm{align} (X)= \omega_\textrm{sparse} E_\textrm{sparse} (X) + \omega_\textrm{dense} E_\textrm{dense} (X)
\end{equation}
where, $\omega_\textrm{sparse}$ and $\omega_\textrm{dense}$ are weights assigned to sparse and dense matching terms and $E_\textrm{sparse}$ (X) and $E_\textrm{dense} (X)$ are the sparse and dense matching terms, respectively, such that: 
\begin{equation}
E_\textrm{sparse} (X) = \sum_\textrm{i=1}^\textrm{$|S|$} \sum_\textrm{j=1}^\textrm{$|S|$} \sum_\textrm{(k,1) $\in$ C(i,j)} ||\tau_iP_\textrm{i,k} - \tau_jP_\textrm{j,k}||^2
\end{equation}
Here, $P_\textrm{i, k}$ is the $k^\textrm{th}$ detected feature point in the $i$-th frame. $C(i,j)$ is the set of all pairwise correspondences between the $i$-th and the $j$-th frame. The Euclidean distance over all the detected feature matches is minimized once the best rigid transformation $\tau_i$ is found. Dense pose estimation is described as follows \cite{dai2017bundlefusion, whelan2015elasticfusion, newcombe2011kinectfusion}:
\begin{equation}
E_\textrm{dense}(\tau) = \omega_\textrm{photo}E_\textrm{photo}(\tau) + \omega_\textrm{geo}E_\textrm{geo}(\tau)
\end{equation}
where,
\begin{equation}
E_\textrm{photo}(X) = \sum_\textrm{(i,j) $\in$ E} \sum_\textrm{k=0}^\textrm{$|I_i|$}|| I_i(\omega(d_\textrm{i,k})) - I_j(\omega(\tau_j^\textrm{-1} \tau_i d_\textrm{i,k}))||_2^2
\end{equation}
and,
\begin{equation}
E_\textrm{geo}(X) = \sum_\textrm{(i,j) $\in$ E} \sum_\textrm{k=0}^\textrm{$|D_i|$} [n^T_\textrm{i,k} (d_\textrm{i,k} - \tau_i^\textrm{-1}\tau_j\omega^\textrm{-1}(D_j(\omega(\tau_j^\textrm{-1}\tau_id_\textrm{i,k}))))]^2
\end{equation}
with $\tau_i$ being rigid camera transformation, $P_\textrm{i,k}$ the $k^\textrm{th}$ detected inlier point in $i^\textrm{th}$ frame, and  $C(i,j)$ being the set of pairwise correspondences between the $i^\textrm{th}$ and $j^\textrm{th}$ frame. In Equation \ref{eq:main}, $\omega_\textrm{dense}$ is linearly increased; this allows the sparse term to first find a good global structure, which is then refined with the dense term (coarse-to-fine alignment \cite{dai2017bundlefusion}). Using Gauss-Newton optimization, we find the best pose parameters $X$ which minimizes the proposed highly non-linear least squares objective.

\subsection{Dataset} 
\label{sec:traindataset_equip}
\begin{figure}[t!]
\centering
\includegraphics[width=\columnwidth]{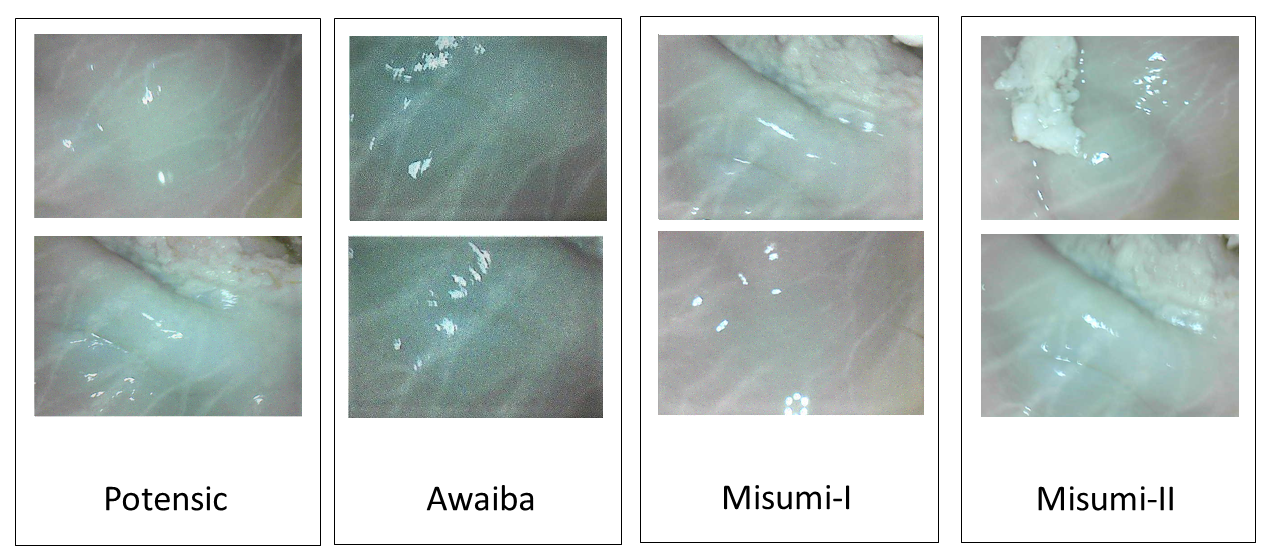}
\caption{Sample frames from the dataset used in the experiments.}
\label{fig:dataset}
\end{figure}

We created our own dataset, which was recorded on five different real pig stomachs. 
To ensure that our algorithm is not tuned to a specific camera model, four different commercial endoscopic cameras were employed. For each pig stomach-camera combination, $2,000$ frames were acquired  which makes for four cameras and five pig stomachs a total of $40,000$ frames. 
Sample images from the dataset are shown in Fig.~\ref{fig:dataset} for visual reference. 
An Optitrack motion tracking system consisting of eight Prime-13 cameras and the manufacturer's tracking software was utilized to obtain 6-DoF pose measurements (see Fig.~\ref{fig:architecture}) as a ground truth for the evaluations of the pose estimation accuracy.
The capsule robot was moved manually with an effort to obtain a large range of poses, during which data was simultaneously recorded from the magnetic localization system, the on-board video camera, and the Optitrack system. We divided our dataset into two groups. A first group consisting of $30,000$ frames was used for RNN training purposes, whereas the remaining $10,000$ frames were used for testing.

\subsection{LSTM Training}
The training data is divided into input sequences of length $50$ and the slices are passed into the LSTM modules with the expectation that it predicts the next 6-DoF pose value, i.e. the $51\text{st}$ pose measurement, which was used to compute the cost function for training.
The LSTM module was trained using Keras library with GPU programming and Theano back-end. Using back-propagation-through-time method, the weights of hidden units were trained for up to $200$ epochs with an initial learning rate of $0.001$.  Overfitting was prevented using dropout and early stopping techniques. Dropout regularization technique introduced by \cite{srivastava2014dropout} is an extremely effective and simple method to avoid overfitting. It samples a part of the whole network and updates its parameters based on the input data. Early stopping is another widely used technique to prevent overfitting of a complex neural network architecture which was optimized by a gradient-based method.

\begin{figure}
\centering
\begin{subfigure}[t]{0.4\textwidth} 
\includegraphics[width=\textwidth]{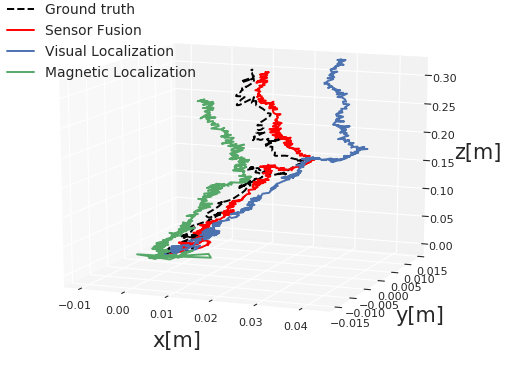}
\caption{Trajectory 1}
\label{fig:traj1} 
\end{subfigure}
~
\begin{subfigure}[t]{0.4\textwidth} 
\includegraphics[width=\textwidth]{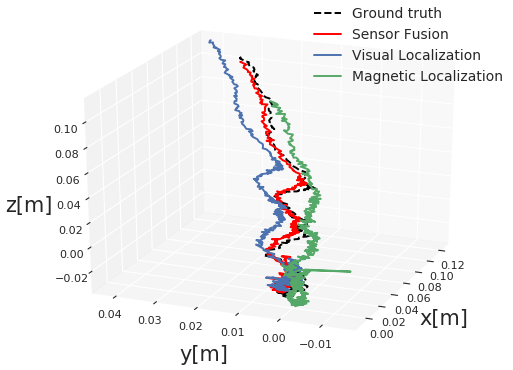}
\caption{Trajectory 2}
\label{fig:traj2} 
\end{subfigure}
~
\begin{subfigure}[t]{0.4\textwidth} 
\includegraphics[width=\textwidth]{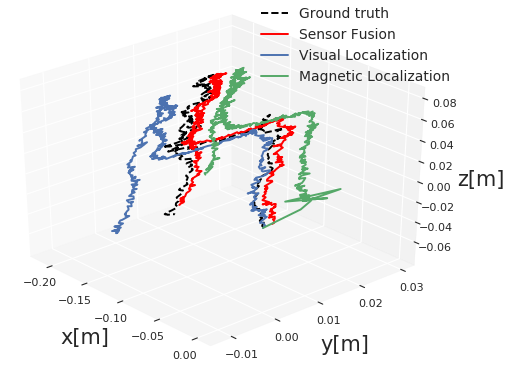}
\caption{Trajectory 3}
\label{fig:traj3}       
\end{subfigure}
~
\begin{subfigure}[t]{0.4\textwidth} 
\includegraphics[width=\textwidth]{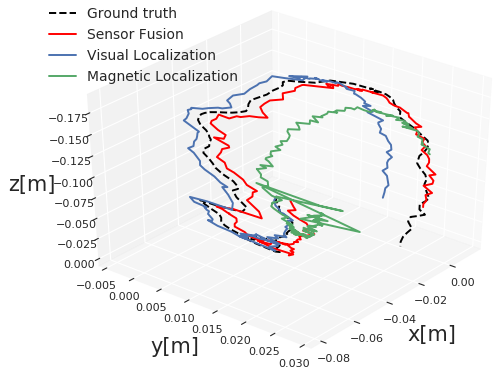}
\caption{Trajectory 4}
\label{fig:traj4} 
\end{subfigure}
~

\caption{Sample trajectories comparing the multi-sensor fusion result with ground truth and sensor data.}
\label{fig:trajs}
\end{figure}

\begin{figure}
  \includegraphics[width=\columnwidth]{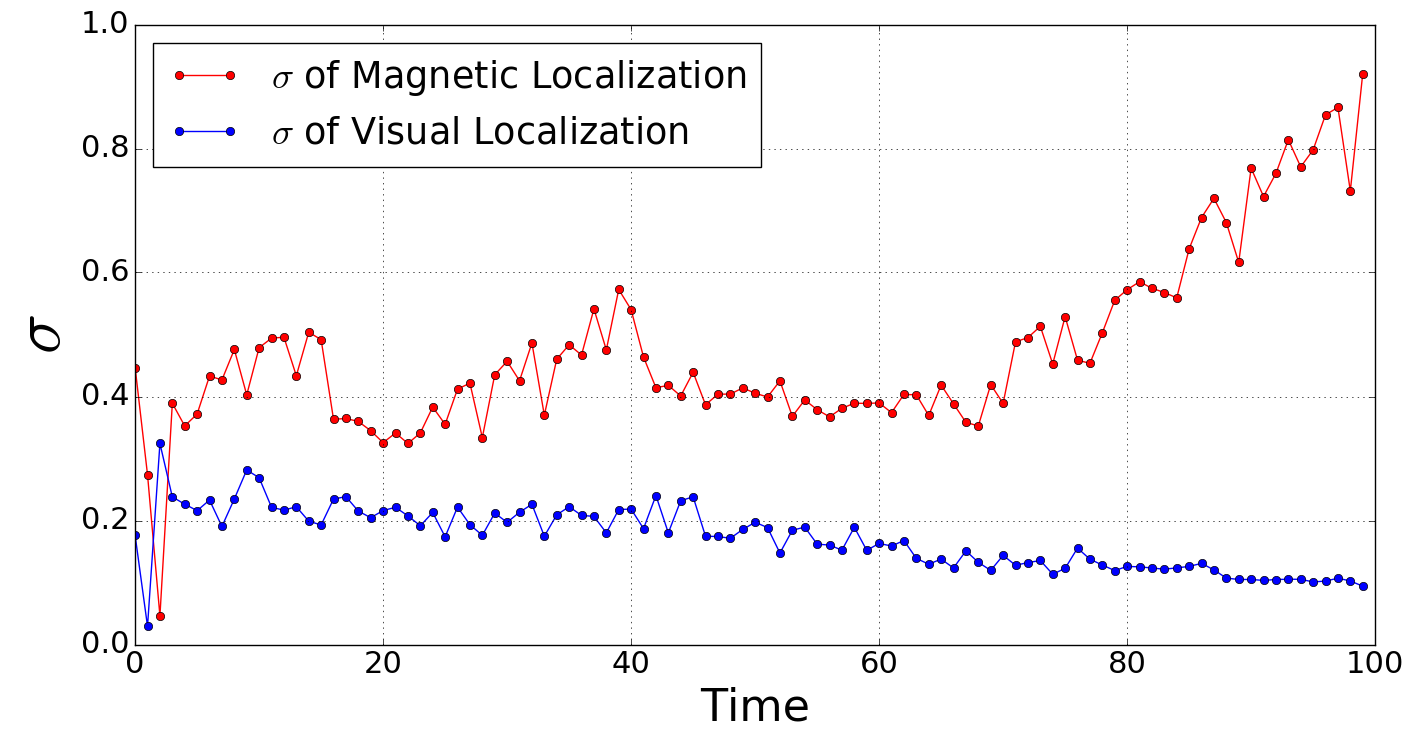}
\caption{Evolution of the $\sigma^{\alpha}_{k,t}$ parameter for the sensors. $\sigma^{\alpha}_{k,t}$ does not tend to increase during sensor failure periods.}
\label{fig:sigma_result}       
\end{figure}

\begin{figure}
\begin{subfigure}{0.95\columnwidth}
\includegraphics[width=\columnwidth]{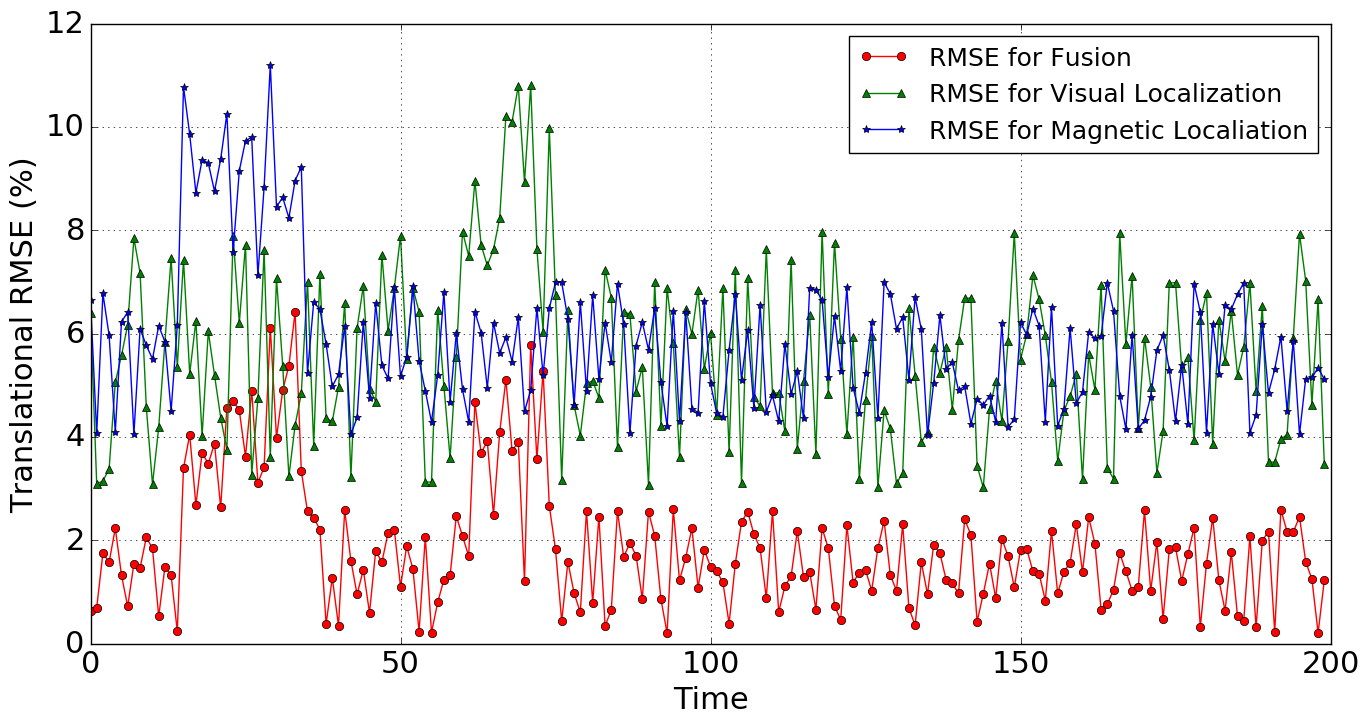}
\label{fig:e1_result}       
\end{subfigure} %
 
\begin{subfigure}{0.95\columnwidth}
\includegraphics[width=\columnwidth]{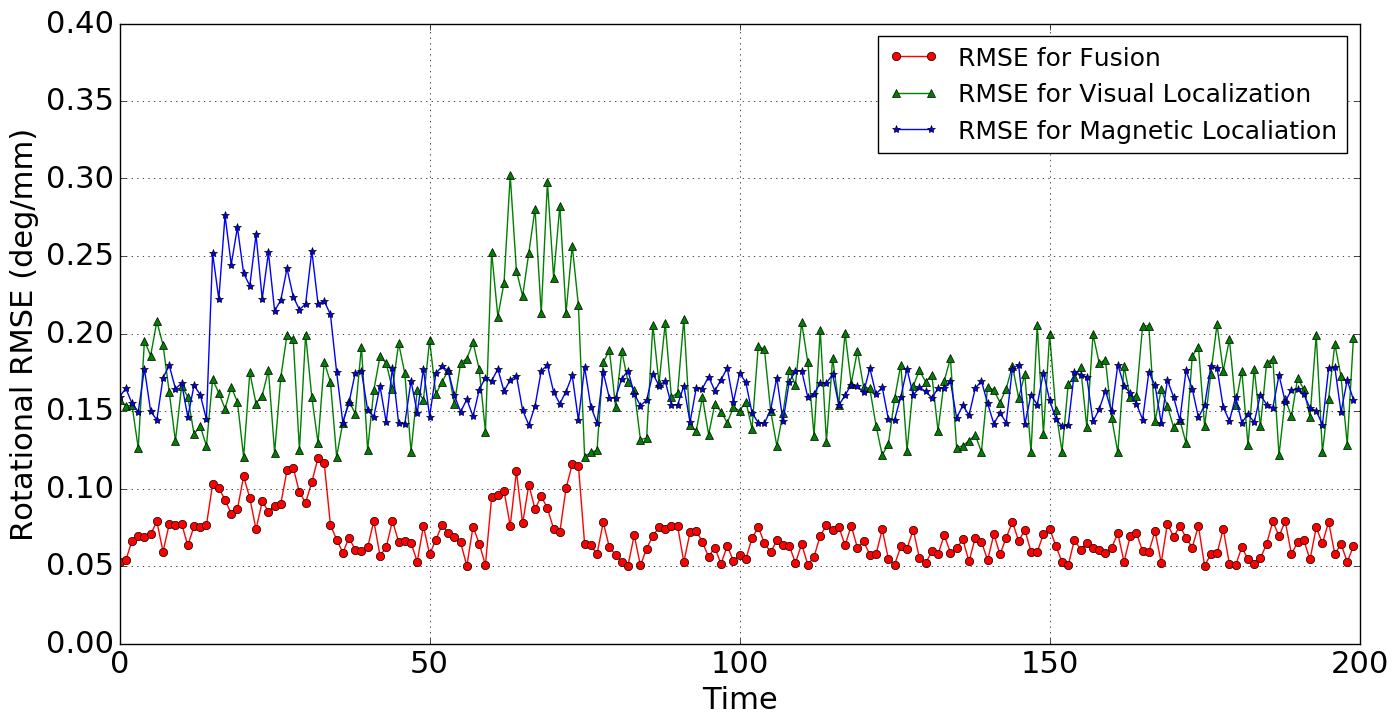}
\label{fig:e2_result}       
\end{subfigure} 
\caption{Translational (top) and rotational (bottom) RMSEs for multi-sensor fusion, visual localization and magnetic localization.}
\label{fig:rmse_result}
\end{figure}  

\section{Results and Discussion}
\label{sec:results_and_discussion}

\begin{figure*}
\centering
\begin{subfigure}[t]{.47\textwidth}
\includegraphics[width=\textwidth]{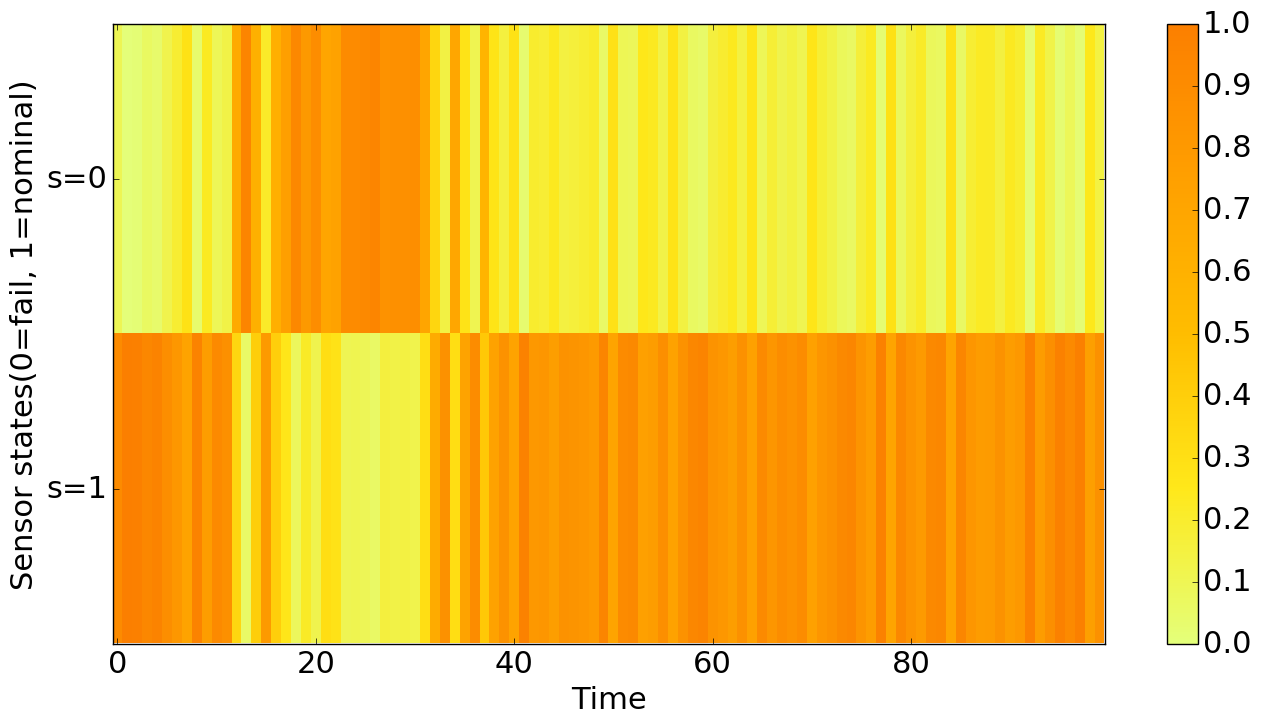}
\label{fig:s1_result}       
\end{subfigure} %
~ 
\begin{subfigure}[t]{0.47\textwidth}
\includegraphics[width=\textwidth]{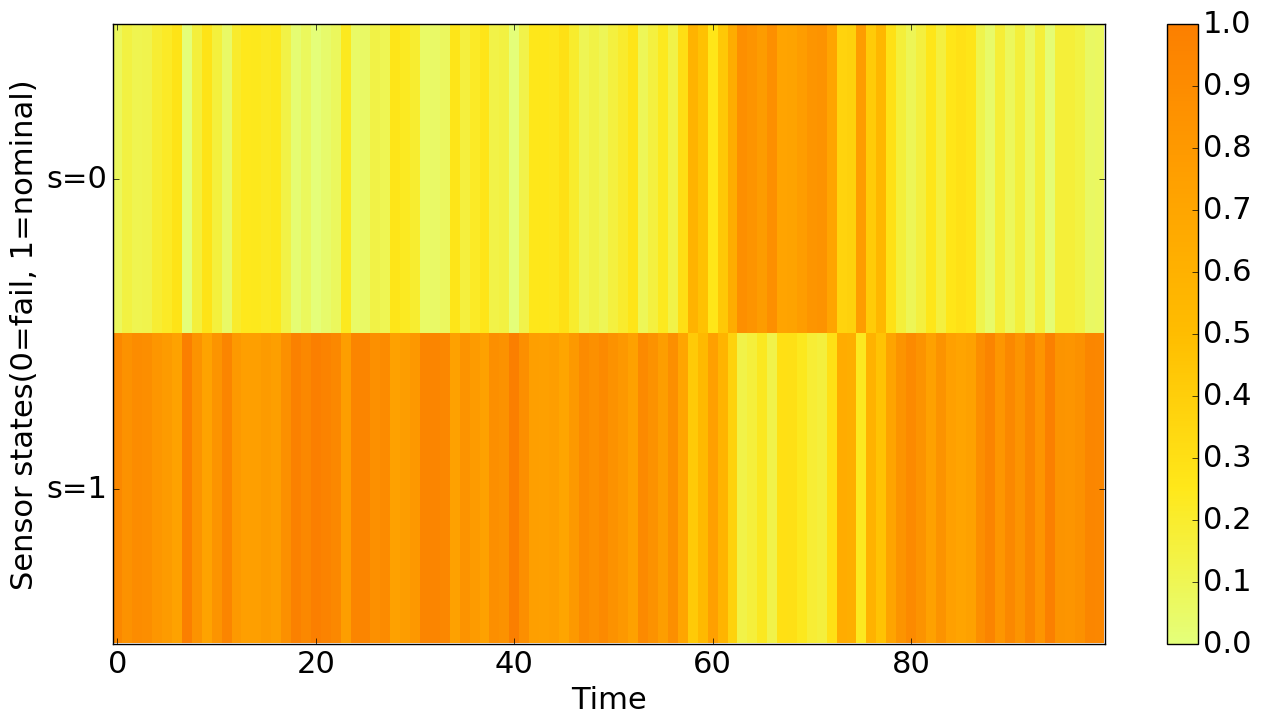}
\label{fig:s2_result}       
\end{subfigure}

\begin{subfigure}[t]{.47\textwidth}
\includegraphics[width=\textwidth]{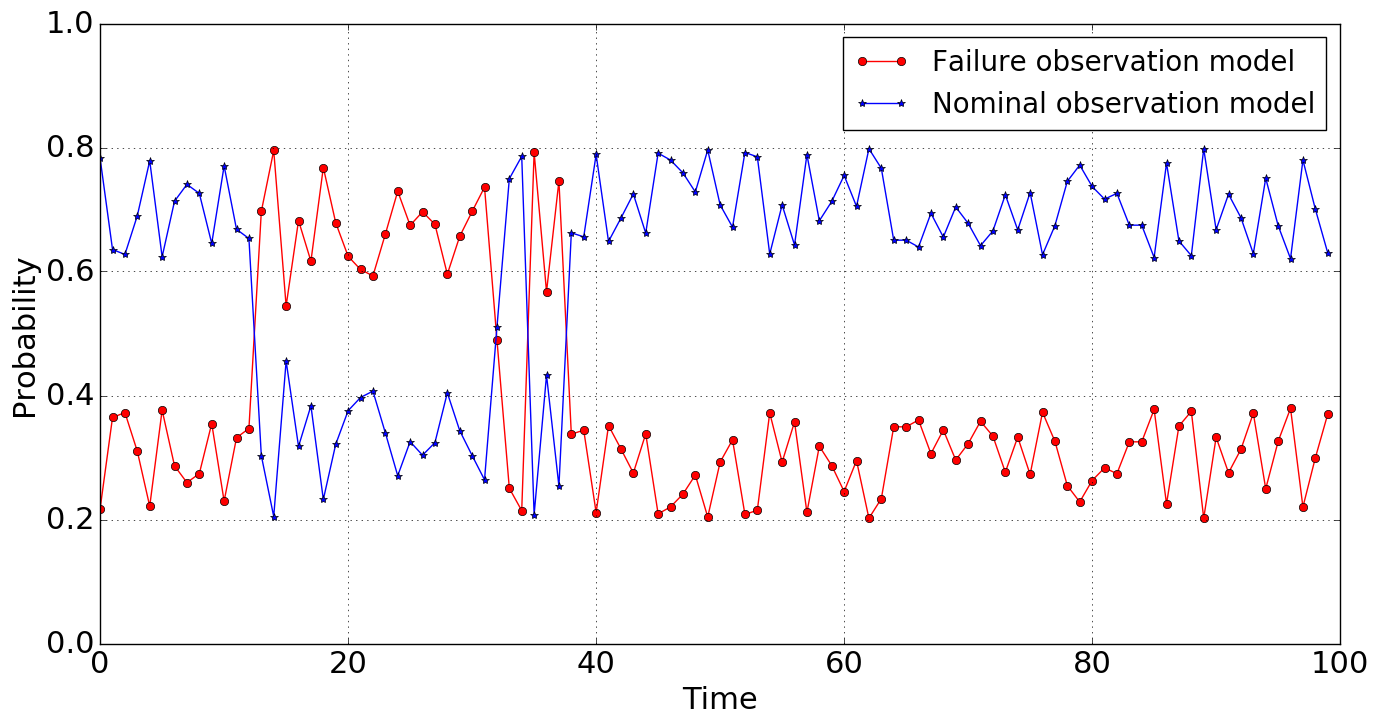}
\label{fig:alpha1_result}       
\end{subfigure} %
~ 
\begin{subfigure}[t]{0.47\textwidth}
\includegraphics[width=\textwidth]{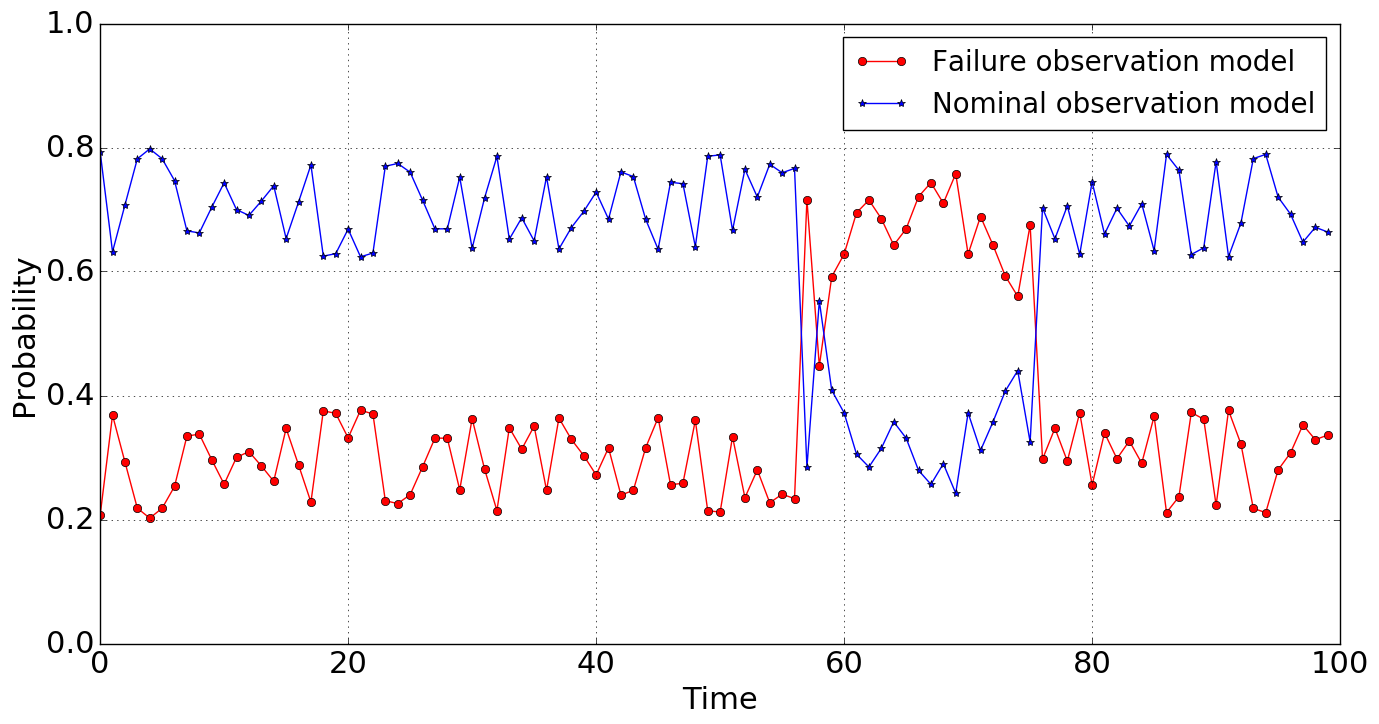}
\label{fig:alpha2_result}       
\end{subfigure}

\caption{Top figures: Posterior probability of $\mathbf{s}_{k,t}$ parameter for endoscopic RGB camera (left) and for magnetic localization system (right). Bottom figures: The minimum mean square error (MMSE) of $\mathbf{\alpha}_{k,t}$ for endoscopic RGB camera (left) and for magnetic localization system (right).  The switch parameter, $\mathbf{s}_{k,t}$, and the confidence parameter $\mathbf{\alpha}_{k,t}$ reflect the failure times accurately: Visual localization fails between 14-36 seconds and magnetic sensor fails between 57-76 seconds. Both failures are detected confidentially.}
\label{fig:alpha_result}
\end{figure*}

The performance of the proposed multi-sensor fusion approach was analysed by examining posterior probabilities of the switch parameters $\mathbf{s}_{k, t}$ (see Fig.~\ref{fig:alpha_result}), the minimum mean square error (MMSE) estimates of $\mathbf{\alpha}_{k,t}$ (see Figure \ref{fig:alpha_result}) and evolution of the hyper-parameter $\sigma^{\alpha}_{k,t}$ (see Fig.~\ref{fig:sigma_result}). Moreover, for various trajectories with different complexity levels of motions, including uncomplicated paths with slow incremental translations and rotations, comprehensive scans with many local loop closures and complex paths with sharp rotational and translational movements, we analysed both the localization accuracy and the fault detection performance of our multi-sensor fusion approach (see Figs.~\ref{fig:trajs} and \ref{fig:rmse_result}). 
Additionally, we compared the rotational and translational motion estimation accuracy of the multi-sensor fusion approach with the visual localization and magnetic localization (see Figure \ref{fig:rmse_result}) using RMSE.

The results in Figure \ref{fig:alpha_result} indicate that  the sensor states are accurately estimated. Visual localization failed because of very fast frame-to-frame motions between 14-36 seconds and magnetic sensor failed due to the increased distance of the ringmagnet to the sensor array between 57-76 seconds. Both failures are detected successfully, and the MMSE is kept low, thanks to the switching option ability  from one observation model to another in case of a sensor failure. In our model, we do not make a Markovian assumption for the switch variable $\mathbf{s}_{k,t}$ but we do for its prior $\mathbf{\alpha}_{k,t}$, resulting in a priori dependent on the past trajectory sections, which is more likely for the incremental endoscopic capsule robot motions. Our model thus introduces a memory over the past sensor states rather than simply considering the last state. The length of the memory is tuned by the hyper-parameters $\sigma^{\alpha}_{k,t}$,  leading to a long memory for large values and vice-versa. This is of particular interest when considering sensor failures. Our system detects automatically failure states. Thus, the confidence in the RGB sensor decrease when visual localization fails recently due to occlusions, fast-frame-to frame changes etc. On the other hand, the confidence in magnetic sensor decreases if the magnetic localization fails due to noise interferences from environment or if the ringmagnet has a big distance to the magnetic sensor array.

The results depicted in Figure \ref{fig:trajs} indicate, that the proposed model clearly outperforms magnetic and visual localization approaches, in terms of translational and rotational pose estimation accuracy. The multi-sensor fusion approach is able to stay close to the ground truth pose values for even sharp crispy motions despite sensor failures. Even for very fast and challenging paths that can be seen in Figure  \ref{fig:traj3} and \ref{fig:traj4}, the deviations of sensor fusion approach from the ground-truth still remain in an acceptable range for medical operations. We presume that the effective use of switching observations and particle filtering with non-linear motion estimation using LSTM enabled learning motion dynamics across time sequences very effectively.


\section{CONCLUSIONS}
\label{sec:conclusion}

In this study, we have presented, to the best of our knowledge, the first particle filter-based multi-sensor data fusion approach with a sensor failure detection and observation switching capability for endoscopic capsule robot localization. 
A LSTM architecture was used for non-linear motion model estimation of the capsule robot. 
The proposed system results in sub-millimetre scale accuracy for translational and sub-degree scale accuracy for rotational motions. Moreover, it clearly  outperforms both visual and magnetic sensors based localization techniques.  As a future step, we consider to integrate a deep learning based noise-variance modelling functionality into our approach to eliminate sensor noise more efficiently.






\bibliographystyle{IEEEtran}
\bibliography{mybibfile}

\end{document}